# Spatiotemporal Emotion Recognition using Deep CNN Based on EEG during Music Listening

**Panayu Keelawat[1], Nattapong Thammasan[2], Masayuki Numao[3], and Boonserm Kijsirikul[1]**

[1]Department of Computer Engineering, Chulalongkorn University, Bangkok, Thailand
[2]Human Media Interaction, University of Twente, Enschede, Netherlands
[3]Institute of Scientific and Industrial Research, Osaka University, Osaka, Japan

Corresponding author: Boonserm Kijsirikul (e-mail: boonserm.k@chula.ac.th).

**ABSTRACT** Emotion recognition based on EEG has become an active research area. As one of the machine learning models, CNN has been utilized to solve diverse problems including issues in this domain. In this work, a study of CNN and its spatiotemporal feature extraction has been conducted in order to explore the model's capabilities in varied window sizes and electrode orders. Our investigation was conducted in subject-independent fashion. Results have shown that temporal information in distinct window sizes significantly affects recognition performance in both 10-fold and leave-one-subject-out cross validation. Spatial information from varying electrode order has modicum effect on classification. SVM classifier depending on spatiotemporal knowledge on the same dataset was previously employed and compared to these empirical results. Even though CNN and SVM have a homologous trend in window size effect, CNN outperformed SVM using leave-one-subject-out cross validation. This could be caused by different extracted features in the elicitation process.

**KEYWORDS** emotion recognition, EEG, machine learning, CNN, spatiotemporal data, brainwave, neuroscience, window size, electrode order

## I. INTRODUCTION

Recently, emotion recognition has become a highly active research area [1]. The reason can be that knowing and understanding emotion is a significant factor in the field of Human-Computer Interaction (HCI). By understanding sentiment, it could manifest a clearer description of human responses to stimuli. Psychologists have studied emotion and endeavored to represent it in a tangible manner. One dominant theory was proposed by Russell [2], indicating that emotion consists of two components, arousal and valence. Arousal designates activation level of emotion, while valence identifies positivity or negativity. This portrayal systematically describes emotions and is widely employed as background knowledge of countless studies, including this current work. Physiological signals are pragmatic tools playing as linkages to emotions, *e.g.*, electrocardiogram (ECG), galvanic skin response (GSR) and electroencephalogram (EEG). There have been many studies to utilize one or multiple signals to conduct research on this topic.

EEG is one way often employed to recognize emotions. As a Brain-Computer Interface (BCI), it is a popular tool frequently used in this task. In many studies, there are numerous techniques which are deployed in order to elicit emotions including images [3], videos [4], or even HCI games [5]. Music is another interesting technique since it can arouse an eclectic range of emotions [6]. In addition, integrating music to emotion recognition based on EEG can produce several useful applications such as music therapy [7] and music recommendation system [8], etc.

In numerous studies, multimodal approach was achieved. Basically, fusion of related information is deployed instead of relying on only a single source. For instance, Verma and Tiwary attempted to identify depression using EEG, GSR, electromyogram (EMG), electrooculogram (EOG), respiratory pattern, blood volume pressure and skin temperature [9]. In addition, brainwave are processed together with features from pupil diameter, gaze distance and eye blinking while watching videos [10]. With respect to music-emotion recognition, there are several studies using multimodal approaches as well. Lin et al. manipulated EEG and musical contents for estimating emotion responses in music listening [11]. Additionally, Thammasan et al. addressed time-varying characteristics of emotion while listening to music combining with amalgamation of EEG and musical features [12]. From our literature review, there are still numerous works we have not mentioned and some research groups that have built multimodal monitoring systems for study convenience [13, 14].

A single-modality approach mostly diminishes emotion recognition performance. Brainwave have extremely low energy and can be susceptible to a great amount of trivial influences. This makes EEG collection oscillate greatly in different experimental settings. However, focusing only on EEG can guarantee that there is no intervention from other non-BCI sources. The study of EEG directly scrutinizes the brain's cognitive state in human mirroring in the collected brainwave and annotated emotion classes.

Conventionally, the task relies on manual feature extraction from recorded signals before feeding in a classifier. Features engineering is normally extracted from band power, signal dimension complexity, etc. Signal complexity along with temporal dimension are some of the most popular features that have been widely used with derivation into fractal dimension [15, 16]. In addition, asymmetry across electrodes has been indicated [11, 17]. It is plausible that spatial and temporal information, or spatiotemporal data, could form into a specific characteristic which identifies brain states. This could be cumbersome for feature extraction. Moreover, all vital features are not warranted to be addressed which can probably hinder optimized performance. With the rise of machine learning, new classification models have been introduced to practically produce leveraged operation on several problems.

Convolutional Neural Network (CNN) has been recently introduced as machine learning. In fact, it has been developed from a simple "neocognitron" [18] to a complex architecture that can solve enigmatic problems nowadays. It has advantages at capturing adjacent spatial information, which, in this case, is spatiotemporal material from EEG. Being an end-to-end model, it has been straightforwardly integrated into a wide variety of applications without manually extracting features. In our previous work on testing in subject-independent manner [19], it showed its potential in generalization to other unseen subjects comparing with another work that operated on Support Vector Machine (SVM) built on Gaussian radical basis kernel using the same dataset [12]. From our review, investigation of CNN gaining advantage from spatiotemporal information is still limited, while there are copious examinations on other classifiers. Although some studies have probed CNN's performance, it is still unclear how it achieves recognition depending on related factors such as window size and electrodes sorting.

In this paper, we aimed to analyze CNN in emotion recognition task based solely on EEG while listening to music along with its effect from electrode order and window size, playing as spatiotemporal inputs. The investigation consisted of multiple network architectures in subject-independent evaluation. According to this analysis, we expected to gain more intellectual insights on CNN and related factors from inputs which possibly impacts its execution. Thus, this intuition can develop a better understanding that furthers the exploration of BCI, neuroscience, cognitive science and other associated subjects in the future.

## II. MATERIALS AND METHODS

### A. DATA COLLECTING AND PREPROCESSING
This part includes several steps from the beginning of the study on which all other following parts were derived.

#### 1) SUBJECTS
Experimental data were acquired from twelve students from Osaka University. All were healthy men with mean age of 25.59 years, standard deviation (SD 1.69 years). Music tracks were used as stimulation and none of the participants ever had formal music education training. Music collection consisted of 40 MIDI files. Each song had only distinct instrument and tempo in order to circumvent influence from lyrics. Subjects were directed to select 16 songs out of the collection. Study participation was voluntary. All subjects were informed about the experimental protocol.

#### 2) DATA COLLECTING PROCEDURE
After selecting 16 songs, the subjects were instructed to listen to them. Songs were synthesized using Java Sound API's MIDI package [20]. There was a 16-second silence interval between songs reducing any effect between songs. While each subject was listening to the selected music track, EEG signals were recorded concurrently. Sampling rate was set at 250 Hz. Signals were collected from twelve electrodes based on a Waveguard EEG cap [21]. We employed the 10-20 international system with Cz as a reference electrode. Those twelve electrodes were all located near frontal lobe that plays an outstanding role in emotion regulation [7], *i.e.*, Fp1, Fp2, F3, F4, F7, F8, Fz, C3, C4, T3, T4 and Pz.

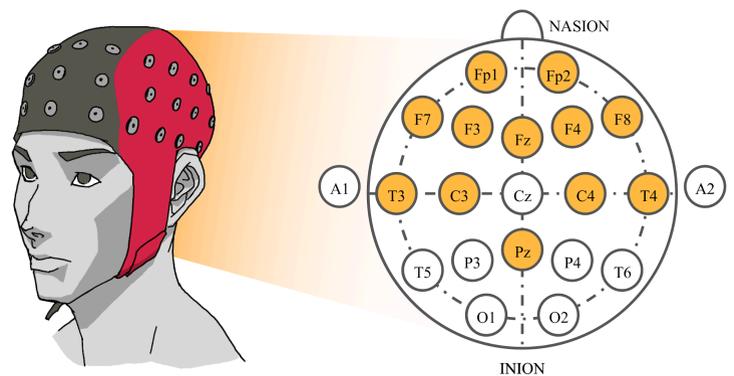

**FIGURE 1.** The 10-20 system of electrode placement showing the selected electrodes.

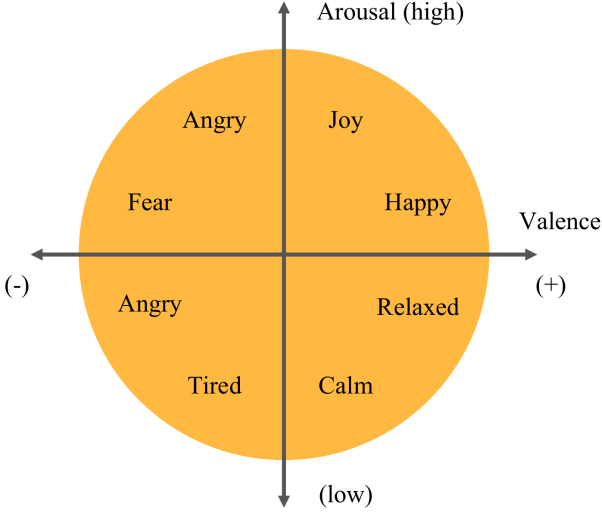

FIGURE 2. The two-dimensional emotion model

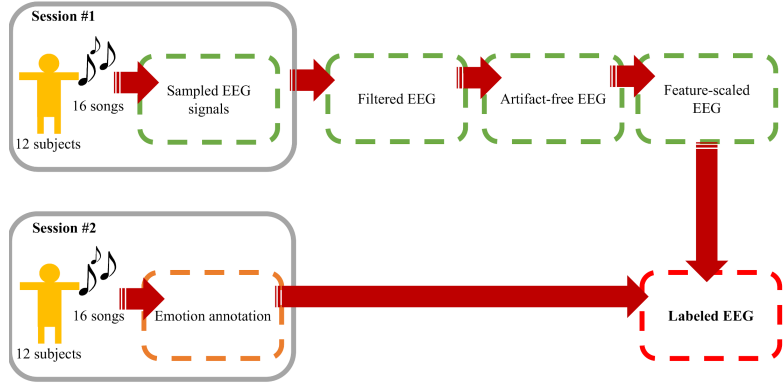

FIGURE 3. Data collecting and preprocessing procedure

Every electrode was adjusted to have impedance of less than 20 kΩ. Sequentially, EEG signals were driven through Polymate API532 amplifier and visualized by APMonitor, both developed by TEAC Corporation [22]. A notch filter was set to the amplifier at 60 Hz. This assured that power line artifact would not interfere with the analysis. Each subject was asked to close his eyes and stay still during the experiment to avoid generation of other unrelated artifacts. When a subject finished the music listening to all 16 songs, the next step was to annotate emotions. The EEG cap was detached. Subjects had to listen to the same songs again in the same order. They labeled their feelings by continuously clicking on a corresponding spot in arousal-valence space displayed on a screen. According to the system, both arousal and valence were recorded separately. Data collection finally finished when data from all subjects were recorded. Next, we applied signal preprocessing.

3) EEG PROCESSING

The frequency range that correlates with this work is 0.5-60 Hz. We employed a bandpass filter to cut off others which were not related. Another tool that was used was EEGLAB [23]. It is an open-source MATLAB environment for EEG processing. We utilized it to remove contaminated artifacts based on Independent Component Analysis (ICA). Those factors could be noise, eye movement, activity from muscle. When every antecedent step was done, signals were associated with emotion annotation via timestamps.

Following, mean and SD were calculated from signal samples, hence we were able to perform feature scaling in terms of standardization. This helps in reducing error rates from high varying signal magnitudes [24].

Labeled emotion elements were categorized into classes. Arousal was divided into high and low, whereas valence was grouped into positive and negative. The emotion model can be viewed as Fig 2. We considered emotion recognition as binary classification task of both components. Data from this process were considered foundation materials for the rest of the experiment in this paper.

Flowchart of data collection and preprocessing is shown below as Fig 3.

B. SUBJECT-INDEPENDENT EXPERIMENT

The attempt to create a universal emotion classifier based on EEG is a challenging topic. Still, there is no standard methodology that gains an unbeatable recognizer for every subject. In this experiment, we considered CNN complexity, window size and electrode ordering as follows.

1) CNN COMPLEXITY

CNN is a broad term that identifies its highlight in computation. Literally, it applies convolution as its base concept. By definition, convolution of discrete function $f$ and $g$, which is denoted by $f * g$, can be calculated as the formula

$$(f * g)[n] = \sum_{-\infty}^{\infty} f[m]g[n - m] \quad (1)$$

The operator can be operated using convolutional layers. We designed and examined four deep CNN models with various number of convolutional layers (described as 3Conv to 6Conv). Techniques such as dropout and max pooling were used for model regularization [25]. After detecting high-level features, every model had fully-connected network for separated arousal and valence prediction at the end. Models can be viewed as Table 1 and Fig 4.

TABLE I
Network Architectures

| Index | 3Conv | | 4Conv | | 5Conv | | 6Conv | |
|---|---|---|---|---|---|---|---|---|
| 1 | Conv 5x5x32 | | Conv 5x5x32 | | Conv 5x5x32 | | Conv 5x5x32 | |
| 2 | Conv 3x3x32 | | Conv 3x3x32 | | Conv 2x2x32 | | Conv 2x2x32 | |
| 3 | MaxPooling 2x2 | | MaxPooling 2x2 | | Conv 2x2x32 | | Conv 2x2x32 | |
| 4 | Conv 3x3x64 | | Conv 2x2x64 | | MaxPooling 2x2 | | MaxPooling 2x2 | |
| 5 | Dropout 0.5 | | Conv 2x2x64 | | Conv 2x2x64 | | Conv 2x2x64 | |
| 6 | FC 128x1 | FC 128x1 | Dropout 0.5 | | Conv 2x2x64 | | Conv 2x2x64 | |
| 7 | Dropout 0.5 | Dropout 0.5 | FC 128x1 | FC 128x1 | Dropout 0.5 | | Conv 2x1x64 | |
| 8 | FC 2x1 | FC 2x1 | Dropout 0.5 | Dropout 0.5 | FC 128x1 | FC 128x1 | Dropout 0.5 | |
| 9 | | | FC 2x1 | FC 2x1 | Dropout 0.5 | Dropout 0.5 | FC 128x1 | FC 128x1 |
| 10 | | | | | FC 2x1 | FC 2x1 | Dropout 0.5 | Dropout 0.5 |
| 11 | | | | | | | FC 2x1 | FC 2x1 |

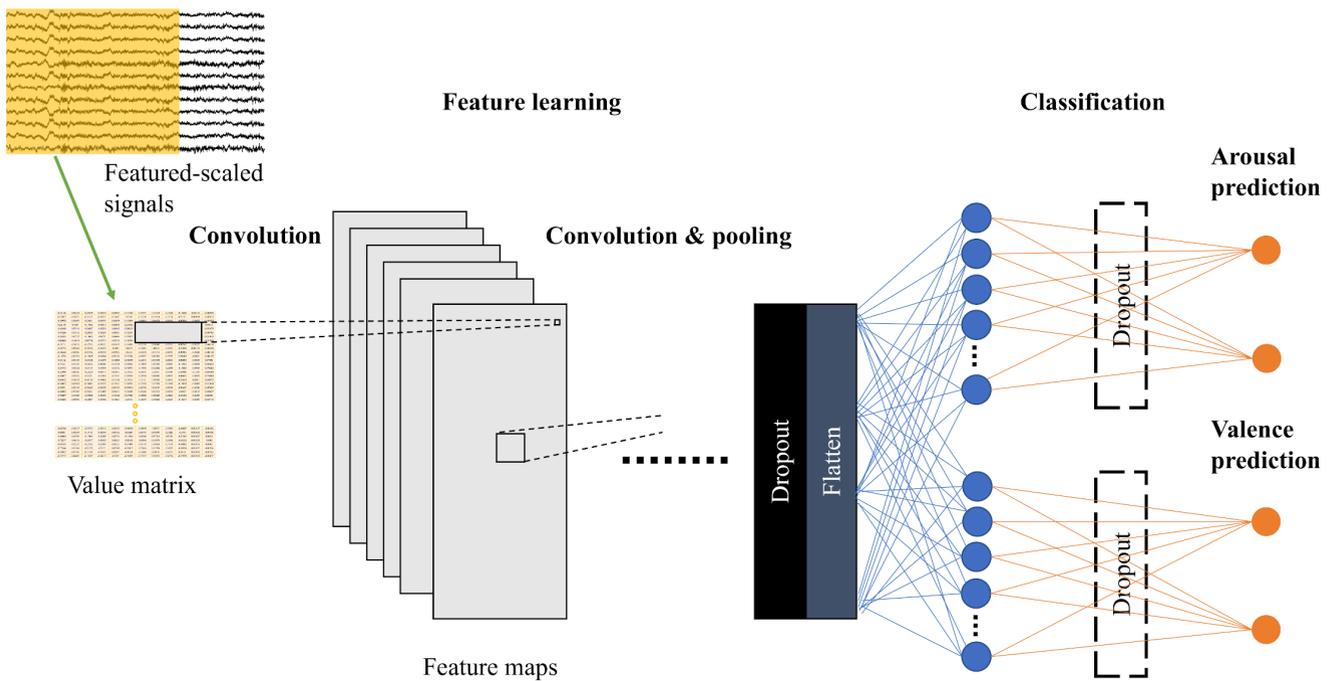

**FIGURE 4. Model illustration**

According to abbreviations, terms are defined as the following. Conv i x j x k means convolutional layer with height of i, width of j and k number of kernels. MaxPooling i x j designates max pooling layer with height of i and width of j. Dropout p conveys the meaning of dropout layer with probability of p being dropped. FC i x j expresses fully-connected layer that has i entries of height and j entries of width. Weights were adjusted by cross-entropy activation function except for the last layer that used softmax. Adam optimizer was adopted for speeding up training [26]. Early stopping was deployed to avoid overfitting. A comparison between models is shown and discussed in subsequent section.

### 2) WINDOW SIZE
Choosing window size is an inevitable part in time series data which include EEG as well. Analyzing these alternatives may be commonplace for traditional machine learning algorithm in this field. Yet, this factor has not been widely addressed on deep CNN architecture. From feature-scaled collected EEG signals, they were divided with time frame in range from 1 to 10 seconds, no overlapping time, and stored in isolated directories. Majority vote was incorporated in this part so that we could earn arousal and valence representation classes for each window. The test consisted of 10-fold cross validation (10-fold CV) covering all subjects and leave-one-subject-out cross validation (LOSO CV). Cross-validation set was randomly selected from training sets of both evaluation methods to avoid overfitting. Evaluation metrics can be referred in "Performance evaluation" section.

### 3) ELECTRODE ORDERING
The most challenging aspect of the brainwave is generalization to unseen subjects. To maximize correlation between subjects, ordering of electrodes is important, especially when employing CNN. The network is good at capturing interconnected spatial information and it has a chance to be affected by juxtaposed electrodes. We designed the order in four types, *i.e.*, randomized order, 3D physical ordering, max adjacent correlation and min adjacent correlation. Results from LOSO CV were compared on window size of 4 seconds, overlapping time 1 second.

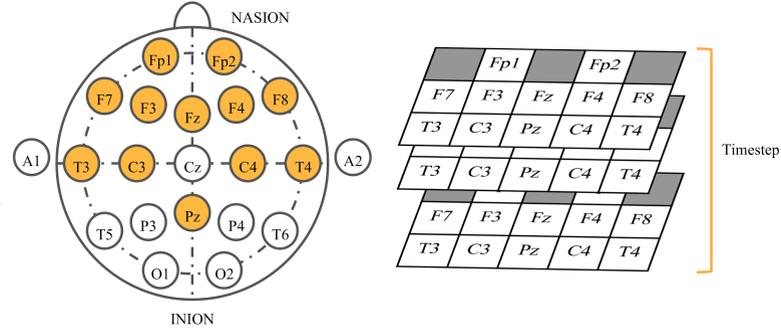

**FIGURE 6. Stacked electrode order regarding physical order on helmet**

From feature-scaled signals, correlation across electrodes could be observed. Some electrodes fetched congruent dynamics which are plotted as sample figures below (Fig 5). Placing electrodes regarding their correlation may perhaps affect the emotion classification.

Sorting by randomizing electrode order was intentionally used as a baseline for other arrangements. We rearranged 20 times and averaged the output performance.

In 3D physical ordering, we mimicked electrode placement on the real device. When gazing from the top view, electrodes are put in a 2D space. They can be represented as a 2D matrix which is shown by Fig 6.

We stacked values placing in this allocation in the order of time steps. Neighbor electrodes may have interrelated correlation possibly increase classification performance. Classifiers were modified to work with 3D inputs.

Next, Pearson Correlation Coefficient (PCC) was calculated from every combination of electrode pairs. PCC is a value that reflects connection between them which can be computed by

$$\rho_{xy} = \frac{\sum_{i=1}^{t}(x_i - \overline{x})(y_i - \overline{y})}{\sqrt{\sum_{i=1}^{t}(x_i - \overline{x})}\sqrt{\sum_{i=1}^{t}(y_i - \overline{y})}} \qquad (2)$$

where $x_i$ represents value of signal sampled from electrode $x$ at time step $i$. Similarly, $y_i$ possesses the same definition for electrode $y$.

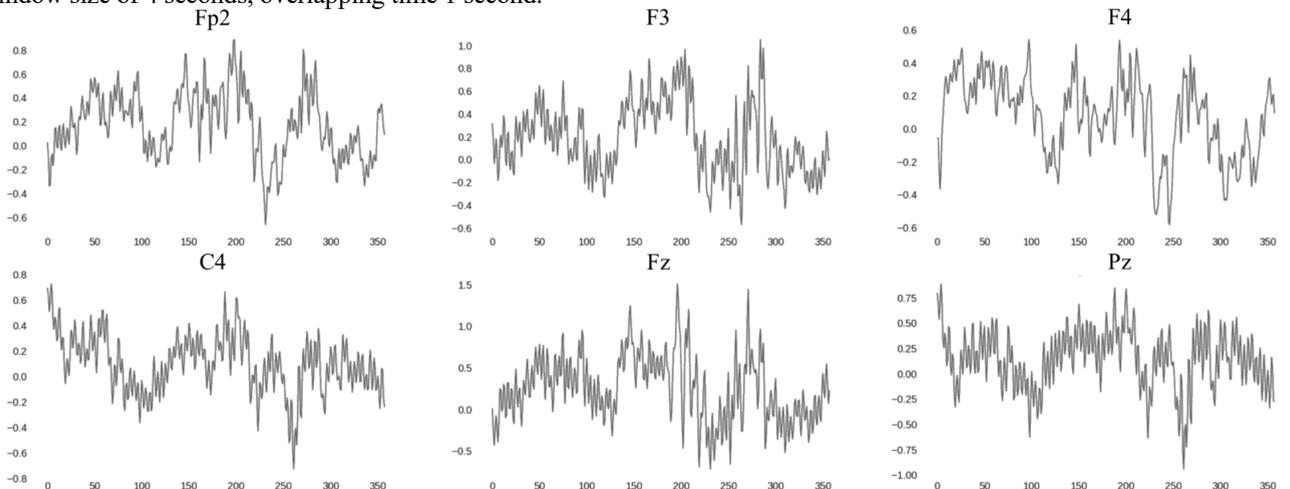

**FIGURE 5. Sample plots of signals**

$t$ is the total samples for electrode $x$ and $y$. Thus, $\rho_{xy}$ denotes PCC between $x$ and $y$. We utilized this information to rearrange electrodes in two algorithms, maximize adjacent pairs [27] and minimize adjacent pairs.

Results from these four methods were evaluated and measured using LOSO CV with fixed window size of 4 seconds and 1 second of overlapping time.

*C. Performance evaluation*

So as to gauge and compare results, sensible metrics should be accounted. As mentioned earlier, we performed two cross validation methods, 10-fold CV and LOSO CV. In both cases, accuracy and Matthews Correlation Coefficient (MCC) [28] were assessed from the separated test set. Accuracy was defined as:

$$Accuracy = \frac{TP + TN}{TP + FP + TN + FN} \times 100 \quad (3)$$

while $TP$, $TN$, $FP$ and $FN$ were denoted as the number of true positives, true negatives, false positives and false negatives, respectively.

Self-reporting emotion state method could lead to produce an imbalance of data. Therefore, we adopted MCC which takes class imbalance factor into account. It can be calculated as:

$$MCC = \frac{TP \times TN - FP \times FN}{\sqrt{(TP + FP)(TP + FN)(TN + FP)(TN + FN)}} \quad (4)$$

while all abbreviations are the same as equation 3. MCC ranges from -1 to 1 inferring all wrong to all correct prediction.

## III. RESULTS AND DISCUSSION

By varying window sizes, we are able to see their effects on performance for both 10-fold CV and LOSO CV. As shown in Table 2, increasing window frame could gain a lower performance in 10-fold CV. This observation applies to every model no matter how complex it is. Also, the outputs from all models seem similar. There is potential that enlarging the frame size can possibly decrease resemblance between instances. In fact, testing the performance by 10-fold CV can produce a test set which has a greater similarity to training set considering that it randomly selects pieces from every subject.

TABLE II
Accuracy and MCC from 10-fold CV Based on Different Window Sizes

| | | Window size (sec) | | | | | | | | | |
|---|---|---|---|---|---|---|---|---|---|---|---|
| | | 1 | 2 | 3 | 4 | 5 | 6 | 7 | 8 | 9 | 10 |
| Arousal | 3Conv | 75.512260 (0.510245) | 66.707170 (0.334143) | 61.569896 (0.231398) | 59.686349 (0.193727) | 60.782640 (0.215653) | 59.510111 (0.190202) | 58.388207 (0.167764) | 59.715554 (0.194311) | 58.031088 (0.160622) | 57.986625 (0.159732) |
| | 4Conv | 77.600002 (0.552000) | 65.754998 (0.315100) | 62.424051 (0.248481) | 61.120324 (0.222406) | 59.939152 (0.198783) | 59.407722 (0.188154) | 59.226454 (0.184529) | 59.810440 (0.196209) | 58.808290 (0.176166) | 58.683963 (0.173679) |
| | 5Conv | 78.360166 (0.567203) | 69.952916 (0.399058) | 61.270763 (0.225415) | 60.537013 (0.210740) | 59.433138 (0.188663) | 58.555873 (0.171117) | 61.619124 (0.232382) | 59.626325 (0.192526) | 59.222798 (0.184456) | 58.969284 (0.179386) |
| | 6Conv | 78.354899 (0.567098) | 65.865875 (0.317317) | 59.449026 (0.188981) | 59.371796 (0.187436) | 59.909242 (0.198185) | 59.816118 (0.196322) | 58.906295 (0.178126) | 60.446776 (0.208936) | 58.290155 (0.165803) | 58.451405 (0.169028) |
| Valence | 3Conv | 81.886288 (0.637726) | 75.371624 (0.507432) | 73.483398 (0.469668) | 73.371355 (0.467427) | 73.172258 (0.463445) | 73.290033 (0.465801) | 73.215299 (0.464306) | 73.392191 (0.467844) | 73.264249 (0.465285) | 73.092150 (0.461843) |
| | 4Conv | 82.822542 (0.656451) | 74.563355 (0.491267) | 73.266174 (0.465323) | 73.414643 (0.468293) | 73.366355 (0.467327) | 73.221657 (0.464433) | 73.058486 (0.461170) | 73.480374 (0.469607) | 72.953368 (0.459067) | 73.088117 (0.461762) |
| | 5Conv | 83.676591 (0.673532) | 75.205413 (0.504108) | 73.099396 (0.461988) | 73.438168 (0.468763) | 73.000396 (0.460008) | 73.391725 (0.467835) | 73.174980 (0.463500) | 73.753299 (0.475066) | 73.264249 (0.465285) | 73.088789 (0.461776) |
| | 6Conv | 82.745467 (0.654909) | 74.429703 (0.488594) | 73.249591 (0.464992) | 73.325508 (0.466510) | 73.309780 (0.466196) | 73.392074 (0.467841) | 73.218327 (0.464367) | 73.614847 (0.472297) | 73.056995 (0.461140) | 72.973854 (0.459477) |

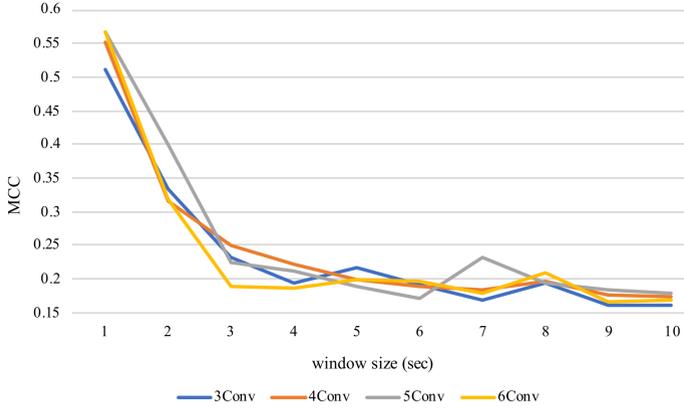

**FIGURE 7. Subject-independent arousal 10-fold CV MCC values with distinct architectures**

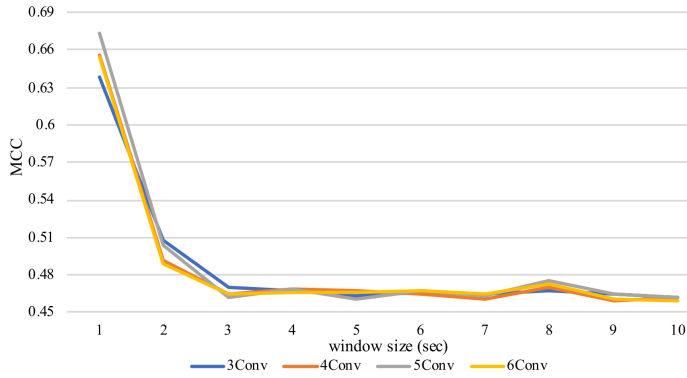

**FIGURE 8. Subject-independent valence 10-fold CV MCC values with distinct architectures and window sizes**

Thus, training covers examples from the whole set of participants and those fragments have higher propensity to be scattered more pervasively throughout the dataset when window size is small. With less parameters to learn, greater spreading in the dataset and higher amount of training examples, using smaller window sizes could increment these factors and generate significantly better results.

On the other hand, evaluating models using LOSO CV is likely to cause adverse outcomes. Even though small window size could affect number of training instances and learning parameters, training and test set were less similar. There could be immense variations between signals from each subject. Expanding window size might help decrease difference among them. In other words, gazing in a larger span of time could reduce fluctuation among instances from distinct subjects. Hence, wider window size may gain a better performance by employing LOSO CV. From Figs 9 and 10, the lines are not increasing steadily but an uprising trend can be found in all models. This also suggests that model complexity does not play a big role affecting the outputs.

TABLE III
Accuracy and MCC from LOSO CV Based on Different Window Sizes

|  |  | Window size (sec) | | | | | | | | | |
|---|---|---|---|---|---|---|---|---|---|---|---|
|  |  | 1 | 2 | 3 | 4 | 5 | 6 | 7 | 8 | 9 | 10 |
| **Arousal** | **3Conv** | 51.973875 (0.039477) | 51.198190 (0.023964) | 53.297832 (0.065957) | 53.829943 (0.076599) | 58.307336 (0.166147) | 54.264346 (0.085287) | 54.697619 (0.093952) | 57.571068 (0.151421) | 54.389915 (0.087798) | 55.961331 (0.119227) |
|  | **4Conv** | 52.253759 (0.045075) | 54.005930 (0.080119) | 56.953875 (0.139078) | 56.216050 (0.124321) | 55.301173 (0.106023) | 56.409596 (0.128192) | 57.518202 (0.150364) | 57.865969 (0.157319) | 57.004601 (0.140092) | 56.456915 (0.129138) |
|  | **5Conv** | 51.585923 (0.031718) | 52.454037 (0.049081) | 50.714366 (0.014287) | 57.806189 (0.156124) | 52.317148 (0.046343) | 56.603117 (0.132062) | 54.610227 (0.092205) | 56.538308 (0.130766) | 57.928869 (0.158577) | 57.517705 (0.150354) |
|  | **6Conv** | 52.545877 (0.050918) | 54.952697 (0.099054) | 58.476066 (0.169521) | 57.349610 (0.146992) | 57.289999 (0.145800) | 58.364474 (0.167289) | 59.384568 (0.187691) | 52.765602 (0.055312) | 57.219417 (0.144388) | 57.447433 (0.148949) |
| **Valence** | **3Conv** | 71.256462 (0.425129) | 70.232125 (0.404643) | 71.664168 (0.433283) | 73.085791 (0.461716) | 73.175037 (0.463501) | 72.929536 (0.458591) | 72.668358 (0.453367) | 73.388377 (0.467768) | 72.286893 (0.445738) | 72.911275 (0.458225) |
|  | **4Conv** | 67.604987 (0.352100) | 70.124416 (0.402488) | 69.604462 (0.392089) | 72.249955 (0.444999) | 72.892736 (0.457855) | 73.046220 (0.460924) | 71.757534 (0.435151) | 73.343332 (0.466867) | 72.599404 (0.451988) | 72.911274 (0.458225) |
|  | **5Conv** | 63.117011 (0.262340) | 67.872853 (0.357457) | 71.758958 (0.435179) | 70.747202 (0.414944) | 73.094099 (0.461882) | 73.149101 (0.462982) | 72.995189 (0.459904) | 73.299005 (0.465980) | 73.019718 (0.460394) | 72.855346 (0.457107) |
|  | **6Conv** | 65.358923 (0.307178) | 71.031241 (0.420625) | 72.682953 (0.453659) | 73.083914 (0.461678) | 72.978358 (0.459567) | 73.043616 (0.460872) | 72.046344 (0.440927) | 73.343332 (0.466867) | 73.019718 (0.460394) | 72.852173 (0.457043) |

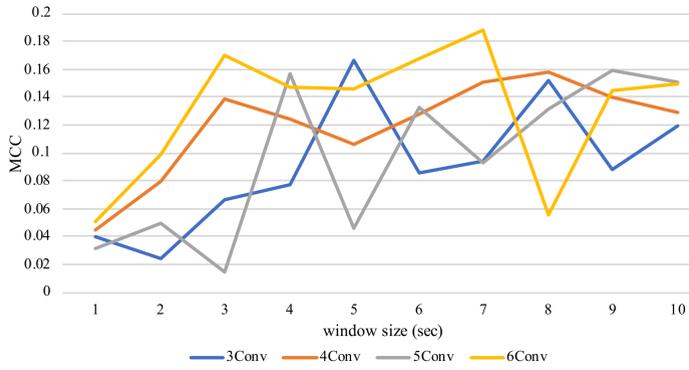

**FIGURE 9.** Subject-independent arousal LOSO CV MCC values with distinct architectures and window sizes

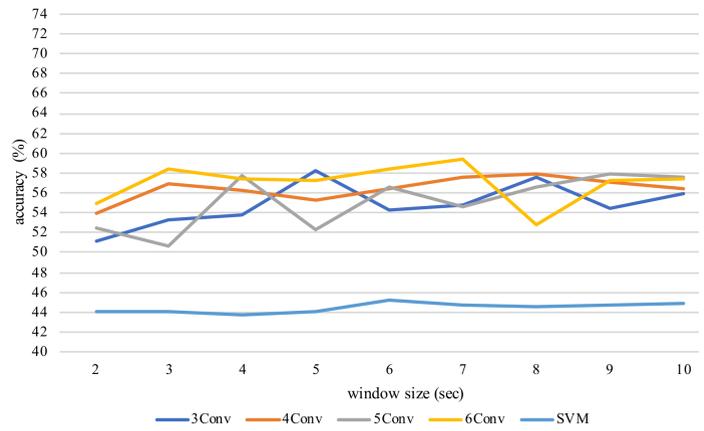

**FIGURE 11.** Subject-independent arousal LOSO CV accuracy of CNNs and SVM

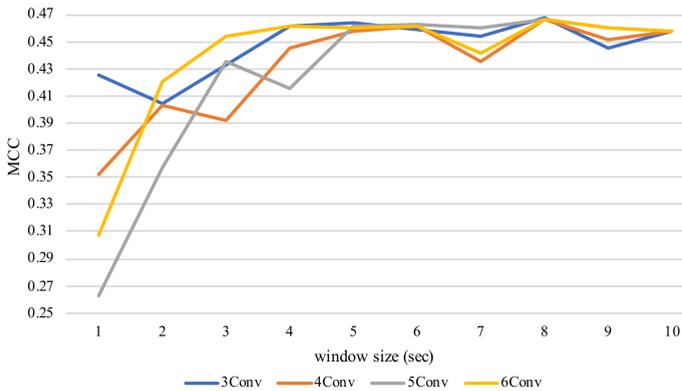

**FIGURE 10.** Subject-independent valence LOSO CV MCC values with distinct architectures and window sizes

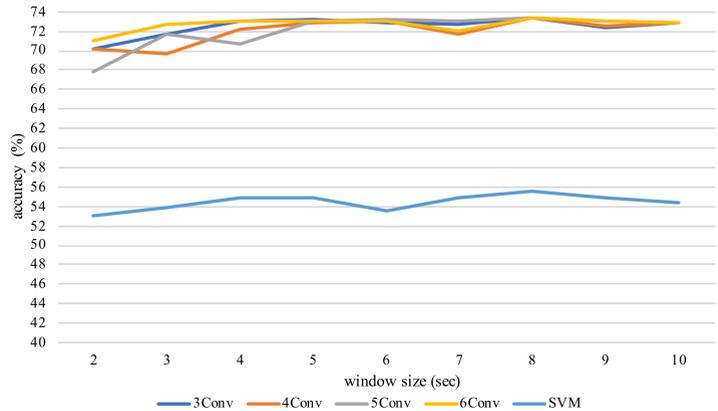

**FIGURE 12.** Subject-independent valence LOSO CV accuracy of CNNs and SVM

The results have similar trends in SVM classifier based on RBF kernel (kernel scale = 3) which had achieved elevation from widening window sizes in LOSO CV. Even though CNN behavior is mostly alike SVM, CNNs are relatively better at recognizing unseen subjects as mirrored from LOSO CV in both valence and arousal. The possible cause can be from the advantage of CNN that utilizes spatiotemporal information to conduct self-governing feature extraction. In other words, while SVM was fed with manually extracted spatiotemporal features, *i.e.*, fractal dimension and asymmetry indexes, CNN excerpted features autonomously. Those extracted attributes may be different from the manually extract ones and may be applicable to various subjects including the unseen ones. Graphs considering results of CNNs together with SVM are illustrated below for patent visualization.

Absolute PCC values were used to decide electrodes sequence using algorithms as described in the previous section. Maximum adjacent PCC was the order of F7, F8, C4, Fz, T4, Pz, T3, C3, Fp2, F3, Fp1 and F4. Besides, minimum adjacent PCC was placed in the order of T4, F3, Pz, Fp1, T3, F4, C3, F7, Fz, F8, C4 and Fp2.

Results from electrode sorting are shown in Table 4. As data suggest, their orders might not induce a significantly high impact on the performance of an unseen subject, although random order never generates the best accomplishment in any architecture. Sometimes the best one from one model is beaten even by the random order scheme in another model.

Table IV
Accuracy and MCC from LOSO CV from Multiple Electrodes Sorting Based on Four Network Architectures

|  |  | Architecture | | | |
|---|---|---|---|---|---|
|  |  | 3Conv | 4Conv | 5Conv | 6Conv |
| Arousal | Random order | 54.5023705 (0.0900475) | 55.7120145 (0.1142405) | 56.0077815 (0.1201555) | 55.9331215 (0.1186625) |
|  | 3D physical placement | 56.411501 (0.128230) | **56.438905 (0.128778)** | 53.791591 (0.075832) | **59.408788 (0.188176)** |
|  | Max adjacent PCC | 56.577301 (0.131546) | 56.156246 (0.123125) | 55.814186 (0.116284) | 55.915050 (0.118301) |
|  | Min adjacent PCC | **56.604240 (0.132085)** | 55.954750 (0.119095) | **57.456453 (0.149129)** | 57.670596 (0.153412) |
| Valence | Random order | 71.8832085 (0.437664) | 72.0806555 (0.441613) | 72.652166 (0.4530435) | 72.9508505 (0.459017) |
|  | 3D physical placement | 72.556503 (0.451130) | **73.060076 (0.461202)** | **73.076878 (0.461538)** | **73.060076 (0.461202)** |
|  | Max adjacent PCC | 71.417263 (0.428345) | 71.021820 (0.420436) | 71.310650 (0.426213) | **73.060076 (0.461202)** |
|  | Min adjacent PCC | **73.042715 (0.460854)** | 72.877022 (0.457540) | 72.011532 (0.440231) | **73.060076 (0.461202)** |

Evaluation method, 10-fold CV or LOSO CV, is a crucial part in affecting inclination of the obtained results due to its subsequent effects such as instances diffusion and their resemblance to the test set. Instance sizes and number of training examples will induce different effects based on evaluation practice which were reflected from the tests we conducted described above. If focusing on generalization performance, then LOSO CV should be more considerable. Nonetheless, if the focus is on universal classifier for known subjects, 10-fold CV could be more appreciable.

By far, CNN has shown its ability to capture spatiotemporal patterns to recognize emotions in unseen subjects. However, results still have limitations and fully cannot address the real reason that makes such outputs, since we have yet to investigate the exact learning the models had undergone. Pinpointing classifier attentive features is an arduous effort. This could be a potential direction for future study. By knowing a deeper understanding of its mechanism, the root cause of its generalization ability can be identified. Also, it may help designing learning networks to be more efficient in this and other similar tasks. Moreover, since CNN has many parameters to learn comparing to SVM on RBF kernel, it may not gain its optimal capability with a limited dataset. Finding more participants with diverse characteristics could also help.

**III. CONCLUSION**
In this paper, we analyzed CNN performance in emotion recognition based solely on EEG during music listening combined with other influential factors. We investigated architecture complexity, window size selection and electrode order. EEG signals were collected from twelve subjects and signal processing techniques were employed to reduce effects from unrelated artifacts. Evaluation of 10-fold CV and LOSO CV was carried out from the preprocessed signals in subject-independent fashion. Results have shown that evaluation method differentiates behavior according to related factors. Enlarging window size reduces accuracy in 10-fold CV which could be caused by increased parameters, instance resemblance and sparsity. Contrarily, wide window size may help in LOSO CV because instances could be less fluctuating across subjects. Furthermore, electrode order is insignificant to the classification.

On a wider scope, CNN outperformed SVM if tested on an unseen subject. This can be observed from LOSO CV. The possible supporting factor is that CNN is capable of using spatiotemporal information to extract features and could be applicable to wider range of participants. Differently, SVM gets better results on 10-fold CV. It can be that the manually extracted features that were employed to serve the classifier are more appropriate to seen subjects rather than features elicited from CNN. Additionally, SVM has less parameters to learn compared to CNN which could be privileged on a limited dataset.

However, there are many issues that are still unclear, namely the exact spatiotemporal features CNN captures in this experiment. Profound analysis on the test instances should be performed as future study in this field. Knowing its mechanism would be easier for researchers to construct a guideline in designing the network more effectively. We could gain better results with more explainable CNN insight.

**ACKNOWLEDGMENT**
We would like to thank Chanokthorn Uerpairojkit for his artwork presented in this paper.

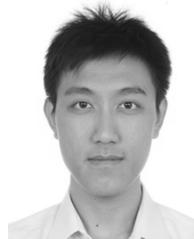

**PANAYU KEELAWAT** has just received B. Computer Eng. degree from Chulalongkorn University, Thailand. In 2018, he was a research intern at Osaka University, Japan. His research interests include machine learning, artificial intelligence, and deep learning.

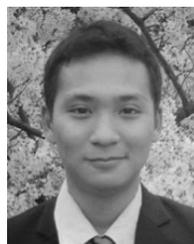

**NATTAPONG THAMMASAN** received the B. Computer Eng. degree in 2012 from Chulalongkorn University, Thailand. He also obtained M. Sc. in 2015 and Ph. D. in 2018 from Osaka University, Japan. He is currently a postdoctoral researcher at Department of Human Media Interaction (HMI), University of Twente, Netherlands. His research interests include affective computing, artificial intelligence, machine learning, and brain-computer interaction.

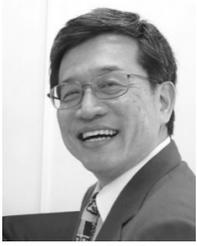
**MASAYUKI NUMAO** is a professor at Department of Architecture for Intelligence, Institute of Scientific and Industrial Research, Osaka University. He received B. Eng. in electrical and electronic engineering in 1982 and his Ph. D. in computer science in 1987 from Tokyo Institute of Technology. He worked at Department of Computer Science, Tokyo Institute of Technology from 1987 to 2003 and was a visiting scholar at CSLI, Stanford University from 1989 to 1990. His research interests include artificial intelligence, machine learning, affective computing, and empathic computing. He is a member of IPSJ, JSAI, and AAAI.

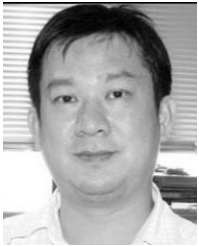
**BOONSERM KIJSIRIKUL** received B. Eng. in electrical and electronic engineering in 1988, M. Eng. in computer engineering in 1990 and D. Eng. in Computer Engineering in 1993 from Tokyo Institute of Technology, Tokyo, Japan. He is currently working as a professor at Department of Computer Engineering, Chulalongkorn University, Bangkok, Thailand. His research interests consist of machine learning, artificial intelligence, natural language processing, etc.